\documentclass[a4paper]{article}

\usepackage{INTERSPEECH_v2}

\usepackage{comment}
\usepackage{amsmath}
\usepackage{amsfonts}
\usepackage{amssymb}
\usepackage{array}
\usepackage{graphicx}
\usepackage{comment}

\newcolumntype{C}[1]{>{\centering}m{#1}}
\newcolumntype{L}[1]{>{\raggedright\let\newline\\\arraybackslash\hspace{0pt}}m{#1}}

\usepackage{cite}

\name{Valentin Barriere$^1$, Chlo\'e Clavel$^1$, Slim Essid$^1$}
\address{
  $^1$LTCI, T\'el\'ecom ParisTech, Universit\'e Paris Saclay, F-75013, Paris, France}
\email{\{firstname.lastname\}@telecom-paristech.fr}

\begin{document}

\title{Opinion Dynamics Modeling for Movie Review Transcripts Classification with Hidden Conditional Random Fields}


\maketitle

\begin{abstract}
In this paper, the main goal is to detect a movie reviewer's opinion using hidden conditional random fields. This model allows us to capture the dynamics of the reviewer's opinion in the transcripts of long unsegmented audio reviews that are analyzed by our system. High level linguistic features are computed at the level of inter-pausal segments. The features include syntactic features, a statistical word embedding model and subjectivity lexicons. The proposed system is evaluated on the ICT-MMMO corpus. We obtain a F1-score of 82\%, which is better than logistic regression and recurrent neural network approaches. We also offer a discussion that sheds some light on the capacity of our system to adapt the word embedding model learned from general written texts data to spoken movie reviews and thus model the dynamics of the opinion.
\end{abstract}



\section{Introduction}
With the growing importance of social networks, the amount of Internet user data has increased dramatically in the last few years. It is now important for companies to exploit this new source of information about their customers in order to be more competitive. The concept of some websites is even to be simply a huge database of recommendations, such as \textit{rottentomatoes.com} where the users rate and review movies, thus delivering their opinion about those movies. 

The domain of opinion mining in textual documents has developed considerably in the last several years. The trend is to use deep learning approaches that allow for achieving high performance, relying on a big amount of training labeled data \cite{Socher2013}. On the other hand, hybrid approaches \cite{yang2013joint} combine the robustness and the high accuracy of Machine Learning (ML) algorithms with the fine-grained modeling of linguistic rules. They do not require a huge amount of labeled data and thus are an interesting alternative to deep learning methods. 

As far as the representation of the data is concerned, various alternatives have been considered in previous works in this area. Using the negations and the intensifiers present in the context of the word as input features for a machine learning algorithm has initially been studied by \cite{Kennedy2006} on the textual IMdB movie review database. The Bag-of-Words (BOW) is a classical domain-agnostic paragraph representation. \cite{Perez-Rosas2013} used BoW and SVM for a sentiment analysis task over the MOUD dataset (Vlogs) obtaining a score of 64.94\%. The Bag-of-N-grams (BoNG), which is an extension of this model, was used by \cite{Schuller2009} for a sentiment analysis task over the Metacritic database (textual movie reviews). \cite{Wagner2014} merge the results of subjectivity lexicons, valence shifters and BoNG to train a classifier for sentiment analysis in tweets. Another trendy option to represent the data nowadays is to create a distributed vector of every word in an unsupervised way, training the model on a large dataset of text. In \cite{Poria2015a}, authors use word2vec with a CNN-SVM for a binary valence classification task on short speech utterances while in \cite{irsoy2014opinion} the authors use the same representation for a task of subjective expression extraction from sentences. In this paper we are in line with all these studies since we combine distributional word embedding with lexicons, linguistic patterns, syntactic features and paralinguistic cues to train a learning model. 

Another issue that is tackled in this paper is how to deal with opinion dynamics in long reviews where the speaker develops his/her opinion across the review. For example, a negative review can include some expressions of positive opinion and then end by a negative opinion. When the size of the documents is increasing, it is crucial to account for the dynamics of the document by using relevant ML method. Opinion dynamics modeling has been rarely addressed in the literature of opinion mining. While some studies are restricted to real-time prediction by segmenting the document into several parts of fixed duration \cite{Poria2016}, others insist on the complementarity of the modalities in order to detect multimodal patterns \cite{morency2011towards,Bousmalis2011}. We distinguish ourselves from these studies by focusing the analysis on the textual modality while using the audio modality only to segment the text based on the pauses of the reviewer. This is motivated by the idea of modeling the opinion dynamics in a more natural way. 

Thus, we investigate a latent state model in order to model the opinion of a speaker along a globally annotated audio movie review. The absence of written punctuation prevent us to segment using syntax so we choose to use oral pauses because of the relevant role of those self-interruptions in the segmentation of discourses\cite{Campione2002}.
Here, we consider the task of labeling an audio transcript with respect to opinions using a variant of Conditional Random Fields (CRF), a discriminative classifier that has proven its utility in several NLP and Computer Vision tasks. This variant, called Hidden Conditional Random Fields (HCRF) has been successfully used to analyze sequences of textual, audio or visual to be labeled globally with only one output \cite{Quattoni2007}. Latent state models have already proven their efficiency for multimodal sentiment analysis or agreement classification \cite{morency2011towards,Bousmalis2011}. 
The objective here is to investigate the potential of HCRF for a classification using transcripts from oral speech. The discriminative nature of CRF will enable some strong linguistic rules combined with other features to emerge directly from the learning phase. 

In the second section of this paper, we will present the features we chose for our task and our learning model. In the third section, we will present the dataset, talk about our experiments and results and finish in the fourth section with a discussion of the results and then we will conclude our paper.

\section{Feature and classification model description}

\subsection{Overview of the system} \label{subsec:overview}

\begin{figure}
\centering
\caption{Overview of the system}
\includegraphics[scale=0.17]{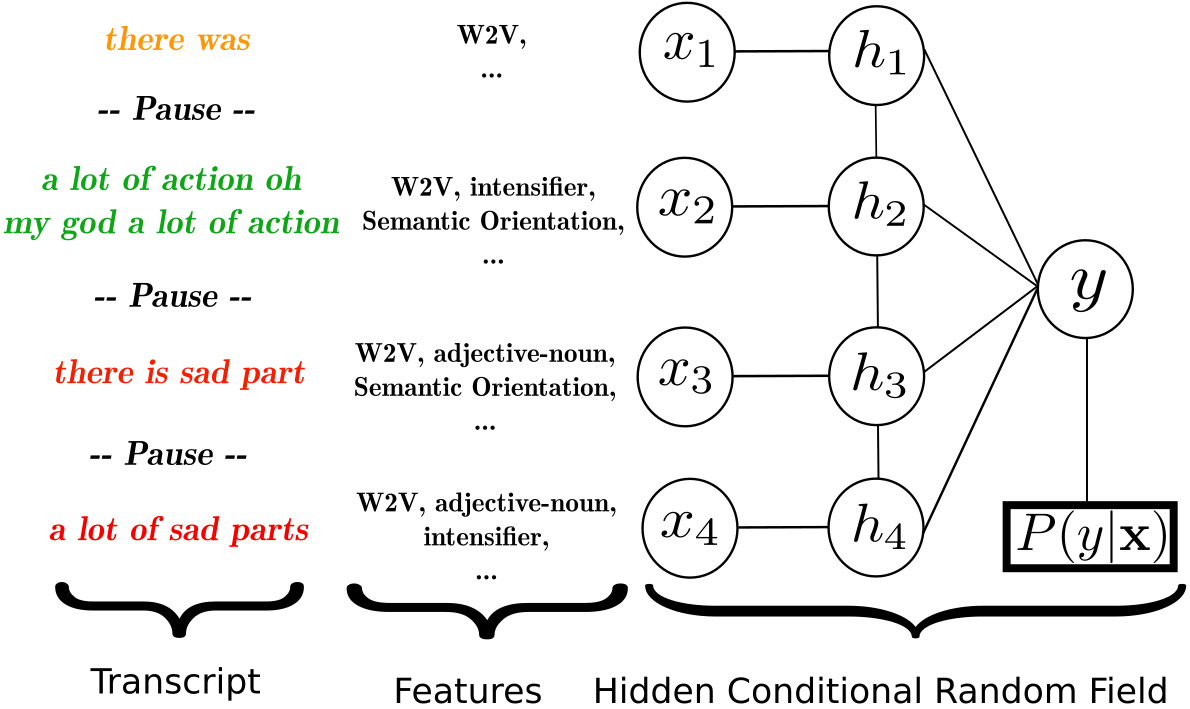}
\label{fig:overview}
\end{figure}

Because of the structure of spontaneous speech, a lot of sentences are unfinished, making it difficult to segment a spoken review into relevant units. We choose to use the pauses to segment the review into Inter Pausal Units (IPUs). Then, we produce the features for each IPU and use them to feed the HCRF, which predicts the most probable label for the current review (see figure \ref{fig:overview}). 

\subsection{Features} \label{subsec:Textual-Features} 
We can sort the textual features we use into 4 groups : 
\\
- \emph{The N-grams features} : The BoNG presented in \cite{Schuller2009} is an extension of the classical Bag of Words representation to N-grams. In this work we use words, bi-grams and tri-grams.
\\
- \emph{The distributed representations} : word2vec is a distributed learning model to represent words \cite{Mikolov2013}. The principle is to use the surrounding words to find the general context in which a word appears and learn its weights statistically. During the learning phase, the vectors of the words appearing in the same context are expected to get closer. This representation can be used to learn more specific semantic information about the discourse of the speaker in the textual features. We chose to use word2vec since it has been found to give better results on a sentiment analysis task in \cite{Poria2015a} compared to other statistical word embeddings. The 300-dimensional vectors we used were pre-trained over a corpus of 100 billions words from Google Press\footnote{Details at https://code.google.com/archive/p/word2vec/}. Generally, it has been empirically found that a more general and bigger training dataset makes it possible to obtain vectors that perform better on several tasks \cite{Mikolov2013a}.
\\
- \emph{The linguistic and lexicon-based features} : The affective valence of a document can be directly retrieved with a rule-based heuristic using specific values attributed for each word with lexicons. We use the negative, positive and neutral SentiWordNet (SWN) scores \cite{Baccianella2010} and the dominance, arousal and valence scores of the enriched Affective norms for English Words (ANEW) lexicon \cite{Warriner2013}. We use linguistic patterns such as adjectives followed by a noun, negations, intensifiers (amplifiers and downtoners). We decided to combine linguistic patterns with sentiment lexicons using the Semantic-Orientation CALculator (SO-CAL) \cite{Taboada2011} which is composed of a grouping of lexicons containing subjective words, intensifiers and valence shifters with associated values. Those values are used for arithmetic operations following simple patterns to give a semantic orientation score to a sentence (see details \cite{Taboada2011}). We separate each value into 3 scores reflecting a positive, a negative and a neutral score so that they can be independently significant of different emotional states of the speaker. 
We finally take the disfluencies, the presence of a capital letter and the 6 parts-of-speech from \cite{Poria2015a} plus interjections and pronouns which are significant of emotional bursts, or belongings.
\\
- \emph{The Paralinguistic features} : The paralinguistic information provided in the transcript can indicate an emotional state which the reviewer does not necessarily evoke through words. The 8 main paralinguistic annotations were dispatched in different categories : the intonation, the pronunciation, the laughter and the volume. 

\subsection{Classification Model}

The HCRF model is used in order to learn a mapping from a sequence of observations $\mathbf{x}_i=\left\{ x_{1},...,x_{L_{i}}\right\}$ of length $L_{i}$ to a label $y_i\in\mathcal{Y}$. Each observation $x_{k}$ is represented by a feature vector $\phi(x_{k})$. For every $\mathbf{x}_{i}$, a sequence of unobserved latent variables $\mathbf{h}_{i}=\left\{ h_{1},...,h_{L_{i}}\right\}$ is defined where $h_{k}\mathcal{\in H}, \mathcal{H}$ being a finite set of states \cite{Quattoni2007}. 
\\
The label decision is made using the posterior probability $P(y|\mathbf{x},\theta)$ given by Eq \eqref{py|x}, where $\theta$ refers to the parameters of the HCRF.  
\begin{equation}
P(y|\mathbf{x},\theta)=\sum_{\mathbf{h}}P(y,\mathbf{h}|\mathbf{x},\theta)=\frac{\sum_{\mathbf{h}}e^{\Psi(y,\mathbf{h},\mathbf{x};\theta)}}{\sum_{y',\mathbf{h}}e^{\Psi(y',\mathbf{h},\mathbf{x};\theta)}},
\label{py|x}
\end{equation}
where $\Psi(y,\mathbf{h},\mathbf{x};\theta)\in\mathbb{R}$ is a potential function (defined in Eq \eqref{Phi}) that measures the compatibility between a label, a sequence of hidden states and the observations. The definition depends on different types of feature functions described below:
\begin{multline}
\Psi(y,\mathbf{h},\mathbf{x};\theta)  = \sum_{j}\langle\phi(x_{j})\,|\,\theta_o(h_{j})\rangle \\ +\sum_{j}\theta_s(y,h_{j})+\sum_{j}\theta_t(y,h_{j},h_{j+1})
\label{Phi}
\end{multline}
- \emph{The hidden state feature functions} depend only on the current observation vector and the current hidden state. A weight $\theta_o(h_{j})$ is created for each hidden state $h_{j}$. The inner product represents the compatibility between an observation and the hidden state.
\\
- \emph{The label feature functions} depend on the label and the current state. The weight $\theta_s(y,h_{j})$ represents the compatibility between a label $y$ and a hidden state $h_{j}$. 
\\
- \emph{The hidden state transition feature functions} depend on the position in the sequence and the label. The weight $\theta_t(y,h_{j},h_{j+1})$ represents the compatibility between a label $y$ and the transition from a hidden state $h_{j}$ to an other hidden state $h_{j+1}$. 
\\
The model is classically trained by minimizing an $\ell_2$-norm regularized negative log-likelihood cost \cite{Quattoni2007}. Decision is taken by choosing the label $y$ that maximizes $P(y|\mathbf{x},\theta)$.

\section{Experiments and results}

We tested three models with different feature sets and segmentations in order to validate our models. Firstly, we created a baseline for our task using a logistic regression model with Bag-of-N-Gram features at document level. 
Since logistic regression does not take into account the dynamics of the observations, we tried a more powerful alternative baseline model that can handle sequential data: a recurrent neural network (RNN-LSTM) \cite{Hochreiter1997}. 
Compared to these models, HCRF offers the benefit of interpretability 
in the way it handles sequential data, while having the potential to model the dynamics of opinion-related phenomena (emotional states, stances, etc.) through latent states. 

Moreover, we compared BoNG to our feature set and tested different pause-based segmentations. 

We used a 10-fold Cross-Validation (CV) where train and test sets are disjoint to validate our models with each test part containing the same proportion of both classes as in the total dataset.

\subsection{Dataset}

In this study, we used the ICT-MMMO corpus\textbf{\footnote{data available by sending an email to abagherz@cs.cmu.edu}} consisting of 365 movie review videos obtained from Youtube.com and ExpoTV.com \cite{wollmer2013youtube}. Those reviews are performed by non-professional users and the audio quality of the recording varies significantly. 
All the videos of the corpus have been annotated in valence by one or two independent annotators. The valence score goes from 1, which means that the speaker has a very negative opinion about the movie, to 5 which denotes a strongly positive opinion about the movie from the speaker, and 3 meaning neutral. The reference is obtained by taking the mean of scores given by the two annotators on a video. The dataset contains more positive videos than negative videos (opinion annotations of the videos : 120 negative, 38 neutral, 207 positive). All the video clips are manually transcribed to extract the spoken words. Using the Transcriber software \cite{Barras2001} each spoken utterance is segmented according to the pause duration. 
All the annotations, the transcriptions of the text and paralinguistic events were made without using the visual information. 

We decided to discard the neutral files because they include files annotated with a different polarity by the two annotators. 
We obtained a total of 321 videos (116 negative and 205 positive) for a total of 13h12 of audio, composed of \textbf{12625} segmented IPUs and \textbf{143181} words.

\subsection{Baselines using LogReg and LSTM}\label{subsec:baseline}

\textbf{\textit{Methodology :}} We considered a baseline model with a simple textual feature set and with our feature set that we tested for different textual representation levels (at the document level or using the pauses) in order to measure the improvement brought by the HCRF. We used logistic regression with a BoNG model like \cite{wollmer2013youtube} with the same parametrization: applying trigram features, Porter stemming, TF-IDF transformations, and document-length normalization. We kept a larger vocabulary. 
We then changed for a more sophisticated feature set (our set in Table \ref{tab:F1-results}), that is a representation using the statistical word embedding model from \cite{Mikolov2013a} described in \ref{subsec:Textual-Features}. After a tokenization\footnote{We used the CoreNLP from Standford \cite{Schuster2016}} we used a spell checker\footnote{https://github.com/phatpiglet/autocorrect} to eliminate the numerous typos from the transcription and to clean the text before taking the word-vectors (stop-words excluded). We followed the protocol of \cite{Mikolov2013a} addressing a sentiment analysis task on short texts and we aggregated by averaging the representations of every word contained inside the IPU to obtain one vector of the same size, and standardized them. In order to help the determination of the opinion we added the linguistic rule set and the values of the subjectivity lexicons (as described in \ref{subsec:Textual-Features}). We used the number of linguistic patterns we detected as well as the scores from the subjectivity lexicons for every word on each IPU to obtain one score per feature on each IPU. We standardized each linguistic feature.
We used pauses longer than 150, 300 and 500 ms (3 experiments) to segment the documents into IPUs.
Regarding the tuning of the logistic regression hyperparameters, we trained with values of the inverse of the regularization strength $C$ in \{0.1, 0.5, 1, 10, 100\}. We used the scikit-learn \cite{Pedregosa2012} implementation of logistic regression. For the RNN-LSTM, we used the keras implementation \cite{Chollet2015} with a number of hidden states in \{64, 128, 256\}, a dropout regularization of $U$ and $W$ (see \cite{Hochreiter1997}) in \{0.1, 0.2, 0.3\} (higher dropout decreased performances) and a number of epochs in \{4...10\}. We used the cross entropy as cost function and Adam as learning algorithm \cite{Kingma2014}.   
\\
\textbf{\textit{Results :}} The results of the baselines are listed in the first part of Table \ref{tab:F1-results} using F1-scores and accuracy. In this table, the global $F1$ (the harmonic mean of recall and precision) is the average $F1$ of both classes ($F1+$ and $F1-$) weighted by their priors and $Accuracy$ is the percentage of true predictions. We notice that the best results are obtained with our feature set. This result is actually unexpected given that we are averaging all the word-vectors of the document into a single one, but the effectiveness comes from the other sentiment-related and linguistic features. The results of the RNN-LSTM are not better for the negative class. Though it has the potential to capture some dynamics, the neural network requires more data than available in the considered corpus to be fully effective. Using the BoNG baseline, \cite{Schuller2009} obtained a F1-score of 78.74\% for a sentiment analysis task 
over the Metacritic database (textual movie reviews).

\subsection{HCRF models}
\textbf{\textit{Methodology :}} 
The existence of latent states in HCRF makes them useful to model a dynamic system like, for example, the emotional state of the speaker. Using our feature set, including sentiment-related features and a distributed representation, the model is expected to more effectively exploit the concepts employed by the speakers. 
We also investigated the granularity of the segmentation, using different thresholds to use the pauses to segment.
We trained the HCRF model with the Matlab wrapper of the HCRF Library \cite{Morency2007} and used a L-BFGS solver for the training. Regarding the exploration of the model hyperparameters, we trained with different values for : the $\ell_2$ regularization parameter in \{0.01, 0.05, 0.075, 0.1, 0.25, 0.5, 1\}, the context window in \{0, 1, 2\} and the number of hidden states in \{2...5\}. The context window is the number of IPU neighbors we concatenated with the centered IPU. We also tested more hidden states, without better results but a longer training time. 
\\
\textit{\textbf{Results :}} The results with the HCRF models are summarized in Table \ref{tab:F1-results}. The best configuration was obtained with 5 hidden states, no context and a value of the regularization parameter equal to 10. As expected, the HCRF improves the results compared to the logistic regression (F1 score improves from 79 to 80) with the BoNG features. The best results were obtained using our set with an improvement of the negative class F1 score (8 points). The 300-ms threshold also brings a slight improvement to the negative class, while using a higher threshold (500 ms) decreases the performance. However, a 10-fold CV does not bring enough information to conclude about the statistical significance of this difference in performance ($p=0.15$ for the negative class).

\begin{table}
\centering
\caption{F1-scores and accuracies results with different feature sets, segmentation thresholds and models}
\label{tab:F1-results}
\begin{tabular}{|L{2.1cm}|ccc|c|c|}
\hline
\textbf{Features} & \textbf{Model} & \textbf{F1+}     & \textbf{F1-}    & \textbf{F1} & \textbf{Acc} \\ \hline \hline
Majority label         & \multicolumn{1}{l|}{Dummy}                  & \multicolumn{1}{l|}{78}         & 0         &  50 & 63           \\ 
BoNG            & \multicolumn{1}{l|}{LogReg}                            & \multicolumn{1}{l|}{84}          & 69          & 78 & 79           \\ 
Our set         & \multicolumn{1}{l|}{LogReg}                            & \multicolumn{1}{l|}{83}          & 72         & 79 & 79           \\ 
Our set (150ms)  & \multicolumn{1}{l|}{LSTM}                            & \multicolumn{1}{l|}{84}          & 68          & 78  & 78          \\ \hline \hline
BoNG (150ms)            & \multicolumn{1}{l|}{HCRF}                      & \multicolumn{1}{l|}{84}          & 67         & 78    & 79        \\ 
Our set (150ms)   & \multicolumn{1}{l|}{HCRF}                      & \multicolumn{1}{l|}{85}          & 72            & 80 & 80         \\ 
Our set (300ms)   & \multicolumn{1}{l|}{HCRF}                        & \multicolumn{1}{l|}{\textbf{86}} & \textbf{75} & \textbf{82} & \textbf{82}  \\ 
Our set (500ms)   & \multicolumn{1}{l|}{HCRF}                      & \multicolumn{1}{l|}{82}          & 67          & 77 & 77          \\ \hline
\end{tabular}
\end{table}

\section{Discussion}

\textit{\textbf{False predictions :}} We provide here an in-depth analysis of false predictions. We found that many examples of wrong classification were due to an opinion too briefly expressed in a review which was globally neutral about the movie. The corresponding videos contain too few linguistic cues of the global opinion of the review. The system also seems to be influenced by the portions of the review where the speaker relates other people's opinion or where they express a strong opinion about something or somebody. There are also cases where the speaker briefly leaves his/her opinions at the beginning or at the end of the video and the main part of the review consists of the reviewer's opinions about general things that are not concerning neither the movie nor its features. Thus, the prediction is complex. The algorithm did not have enough examples to check that the most important points are the opinions of the speaker related to the movie and its features. 
Finally, when examining the pause segmentation ground-truth, we see some segmentation errors: there are 108 IPUs containing more than 50 words (using the 150-ms threshold). Besides, errors are dispatched over more than 18\% of the files of the corpus. A clean pause detection method based on a text aligner could be an effective solution to this problem. 
\\
\textit{\textbf{Hidden states, transitions and activation words :}} After each training of a HCRF model, there is, for each label $y$ at least one state $h_{y}$ with compatibility weight $\theta_s(y,h_{y})$ that is highly positive with one label and highly negative with the other label. The transition between those states is highly improbable. We will call those states '\textit{negative state}' (\textit{Neg}) and '\textit{positive state}' (\textit{Pos}) even if it is an abuse of language. The three other states are considered 'neutral' (\textit{Neu1}, \textit{Neu2}, \textit{Neu3}), with low amplitude transition and compatibility weights. Those states can be used as a bridge between positive and negative states to model the development of the opinion dynamics. 
In Table \ref{tab:features_states_pros}, we present the most relevant examples of the most compatible features with each hidden state (features that correspond to the 30 highest positive weights). In the first column, we can see that the linguistic and paralinguistic features have a less important weight in the neutral states: the only feature having a positive weight for all the neutral states is '\textit{*chuckling*}' while \textit{Pos} and \textit{Neg} have numerous and various linguistic and paralinguistic features with high positive weights. 

\begin{table}
\centering
\caption{Most relevant examples of features with high positive values for each state (paralinguistic with *)}
\label{tab:features_states_pros}
\begin{tabular}{C{0.6cm}|L{3.8cm}|L{3cm}}
\textbf{States} & \textbf{Linguistic and paralinguistic features} & \textbf{Words corresponding to compatible vectors} \\ \hline
\textit{Pos} & adj, disfluency, conjunction, intensifier, *lip smacking*, ...    &  \textit{honors}, \textit{fearless},  \textit{awesome}, \textit{fantastic}  \\
\textit{Neu1} & *chuckling*  & \textit{um}, \textit{Uh}, \textit{ah}, \textit{dunno}, \textit{nada}         \\
\textit{Neu2} & *chuckling*    & \textit{um}, \textit{Uh}, \textit{ah}, \textit{dunno}, \textit{nada}         \\
\textit{Neu3} & $\varnothing$    & \textit{Thanks}, \textit{justin}, \textit{sean}, \textit{michael}, \textit{Sorry}       \\
\textit{Neg} & negation, *falling intonation*, interjection, *word elongation*, ...    & \textit{miserably}, \textit{disappointing}, \textit{yelling}, \textit{failure}, \textit{lack}           \\
\end{tabular}
\end{table}

Regarding word embedding features, our system is no longer learning words but concepts in the 300-dimensional word2vec space by using the information contained inside the word-vectors. In order to analyze the features of the word2vec space, we look for vectors of the words contained in our corpus that activated each state the most. In the second column of Table \ref{tab:features_states_pros}, we can see activation words with high valences, e.g. '\textit{disappointing}', '\textit{miserably}' and '\textit{awesome}'. 
\begin{table}
\centering
\caption{Examples of differences in feature function values}
\label{tab:diff_word_vectors}
\begin{tabular}{|c|c|c|c|}
\textbf{Words} & \textbf{Positive State} & \textbf{Neutral States} & \textbf{Negative State} \\ \hline
\textit{Uh} & -2.57 & 2.8& -3.86  \\
\textit{Yeah} & 0.98 & 2.21& -5.49  \\
\textit{Yes} & -0.06 & 1.21& -3.14  \\
\textit{Thanks} & 2.25 & 3.48& -7.41  \\
\end{tabular}
\end{table}
\\
\textit{\textbf{Role of neutral states :}} 
Learning word embeddings requires a significant amount of text data to be available, that is the reason why we choose to use pre-trained word embeddings. It is interesting to notice that, even though the used word-vectors were learned from general text data, they include spontaneous speech words, such as '\emph{uhm}' or '\textit{dunno}'. However, they do not correspond to the ones that would have been learned on an audio monologue such as the reviews analyzed in this work. For example, while a written '\emph{uhm}' in a post may be a stylistic effect aiming at sounding negative, the oral counterpart is a common hesitation and is possibly neutral. Another example is the difference between  \emph{yes} and \emph{yeah}: the latter is not common in written text where it reflects a more positive thought (see Table \ref{tab:diff_word_vectors}). Further, some other words are merely corpus-specific, e.g. \textit{Hi} and \textit{Thanks} (“\textit{Thanks for watching me guys}”) but associated with positive valence by their word2vec trained on text data. Consequently, information inside the word-vectors may sometimes not be adapted to the discourse of the speaker. The hidden neutral states of the HCRF seem to be handling this issue, so that the problematic word-vectors do not affect the states linked with the global labels of the review.

\section{Conclusion and future work}

In this paper, we have presented a HCRF model that uses a pause-based segmentation of movie review transcripts in order to model the dynamics of the opinion of the speaker through latent states. Our textual feature set includes word embedding, linguistic rules and clues from subjectivity lexicon. The use of HCRF classifiers allows us to implicitly learn local linguistic representations of each inter-pausal segment of the reviews making the integration of word embeddings in the classification system more meaningful. 
We also investigated a pause-based segmentation on a long unannotated discourse, finding that too long segments lead to a loss of performance. 
\\
In our future work we would like to improve the way we use the word embedding in our model in order to handle more precise concepts with more hidden states. 
Further, we would like to test on a bigger corpus in order to obtain significant results.



\bibliographystyle{plain}
\bibliography{main.bib}

\end{document}